\documentclass[lettersize,journal,twoside]{IEEEtran}
\usepackage{amsmath,amsfonts}
\usepackage[linesnumbered,ruled,vlined]{algorithm2e}
\usepackage{array}
\usepackage[caption=false,font=footnotesize,labelfont=rm,textfont=rm]{subfig}
\usepackage{textcomp}
\usepackage{stfloats}
\usepackage{url}
\usepackage{verbatim}
\usepackage{graphicx}
\usepackage{cite}
\usepackage{amssymb}
\usepackage[hidelinks]{hyperref}
\usepackage{multirow}
\usepackage{booktabs}
\usepackage{placeins}
\usepackage{fancyhdr}

\begin{document}

\title{Omni-LIVO: Robust RGB-Colored Multi-Camera Visual-Inertial-LiDAR Odometry via Photometric Migration and ESIKF Fusion}

\author{Yinong Cao$^{1}$, Chenyang Zhang$^{1}$, Xin He$^{1,*}$, Yuwei Chen$^{1}$, Chengyu Pu$^{1}$, Bingtao Wang$^{1}$, \\ 
Kaile Wu$^{1}$, Shouzheng Zhu$^{1}$, Fei Han$^{2}$, Shijie Liu$^{1}$, Chunlai Li$^{1}$ and Jianyu Wang$^{1,3}$
% <-this % stops a space
\thanks{Manuscript received: September 2, 2025; Revised: November 9, 2025; Accepted: January 13, 2026.}% <-this % stops a space

\thanks{This paper was recommended for publication by Editor Pascal Vasseur upon evaluation of the Associate Editor and Reviewers comments.}% <-this % stops a space

\thanks{This work was supported by the Zhejiang Provincial ``Jianbing Lingyan'' R\&D Program under Grants 2024C01126, 2023C03012, and 2025002039, and by Hangzhou Institute for Advanced Study, UCAS, under Grants B02006C021026, B02006C021022, and B02006C021029.}% <-this % stops a space

\thanks{$^{*}$Corresponding author: Xin He (e-mail: xinhe@ucas.ac.cn)}% <-this % stops a space

\thanks{$^{1}$Hangzhou Institute for Advanced Study, University of Chinese Academy of Sciences, Hangzhou 310024, China.}% <-this % stops a space
\thanks{$^{2}$Advanced Laser Technology Lab of Anhui Province, Hefei 230026, China.}% <-this % stops a space
\thanks{$^{3}$Key Laboratory of Space Active Opto-Electronics Technology, Shanghai Institute of Technical Physics, Chinese Academy of Sciences, Shanghai 200083, China}% <-this % stops a space

\thanks{Digital Object Identifier (DOI): see top of this page.}% <-this % stops a space
}

% The paper headers - must be defined before \maketitle
\fancypagestyle{plain}{%
\fancyhf{}
\fancyhead[LO]{\footnotesize IEEE ROBOTICS AND AUTOMATION LETTERS. PREPRINT VERSION. ACCEPTED JANUARY, 2026}
\fancyhead[RE]{\footnotesize IEEE ROBOTICS AND AUTOMATION LETTERS. PREPRINT VERSION. ACCEPTED JANUARY, 2026}
\fancyfoot[C]{\thepage}
\renewcommand{\headrulewidth}{0pt}
}
\pagestyle{fancy}
\fancyhf{}
\fancyhead[RE]{\footnotesize IEEE ROBOTICS AND AUTOMATION LETTERS. PREPRINT VERSION. ACCEPTED JANUARY, 2026}
\fancyhead[LO]{\footnotesize Cao et al.: Omni-LIVO: Robust RGB-Colored Multi-Camera Visual-Inertial-LiDAR Odometry via Photometric Migration and ESIKF Fusion}
\fancyfoot[C]{\thepage}
\renewcommand{\headrulewidth}{0pt}

\maketitle
\thispagestyle{plain}

\begin{abstract}
Wide field-of-view (FoV) LiDAR sensors provide dense geometry across large environments, but existing LiDAR-inertial-visual odometry (LIVO) systems generally rely on a single camera, limiting their ability to fully exploit LiDAR-derived depth for photometric alignment and scene colorization. We present Omni-LIVO, a tightly coupled multi-camera LIVO system that leverages multi-view observations to comprehensively utilize LiDAR geometric information across extended spatial regions. Omni-LIVO introduces a Cross-View direct alignment strategy that maintains photometric consistency across non-overlapping views, and extends the Error-State Iterated Kalman Filter (ESIKF) with multi-view updates and adaptive covariance. The system is evaluated on public benchmarks and our custom dataset, showing improved accuracy and robustness over state-of-the-art LIVO, LIO, and visual-inertial SLAM baselines.
\end{abstract}


\section{Introduction}
\IEEEPARstart{S}{imultaneous} Localization and Mapping (SLAM) has become essential for autonomous systems operating in unknown environments. Traditional methods are categorized by sensor types: vision-based SLAM \cite{engel2014lsd,mur2017orb} provides rich visual information but struggles in low-light or texture-poor scenes, while LiDAR-based SLAM \cite{zhang2014loam} offers robust geometry but degrades in repetitive environments. IMUs, though environment-independent, suffer from long-term drift.

To address single-sensor limitations, multi-sensor fusion systems like visual-inertial SLAM (VI-SLAM) and LiDAR-inertial SLAM (LI-SLAM) enhance robustness through complementary sensor properties \cite{qin2018vins,campos2021orb}. Recent LIVO systems combine dense LiDAR geometry, visual measurements, and high-frequency IMU measurements. LVI-SAM \cite{lvisam2021shan} employs feature-based visual tracking, while FAST-LIVO \cite{zheng2022fast} and FAST-LIVO2 \cite{zheng2024fast} achieve high accuracy through photometric patch alignment on LiDAR-derived map points.

Multi-camera configurations offer significant advantages for LIVO systems through extended spatial coverage, enabling comprehensive utilization of LiDAR-derived depth for photometric alignment, richer geometric constraints, improved observability, and enhanced robustness under occlusion and illumination changes. While multi-camera systems have been explored in visual SLAM \cite{kuo2020redesigning,wang2024mavis}, their tight integration with LiDAR and inertial data in LIVO systems remains underexplored. 

Realizing these advantages requires addressing key technical challenges. Unlike stereo systems that use overlapping views for feature matching, non-overlapping multi-camera configurations require different approaches. First, photometric continuity must be maintained through temporal patch migration as patches transition between cameras. Second, adaptive covariance modeling is needed to handle heterogeneous measurement reliability across different views \cite{zhang2022multicam,fontan2022covariance}. Directly leveraging LiDAR-derived depth for multi-camera sparse direct alignment maintains efficient performance while enabling accurate visual point selection and extended spatial coverage beyond single-camera constraints.

% \IEEEpubidadjcol

To address these challenges, we present Omni-LIVO\footnote{Code and data are available at \url{https://github.com/elon876/Omni-LIVO}.}, a tightly coupled multi-camera LIVO system that extends FAST-LIVO2 with multi-view photometric constraints.

The main contributions are:
\begin{enumerate}
\item \textbf{Cross-View temporal migration}: A direct alignment strategy preserving photometric consistency across non-overlapping views.
\item \textbf{Adaptive multi-view ESIKF}: Enhanced ESIKF with adaptive covariance for improved robustness in adverse conditions.
\item \textbf{Extensive experimental evaluation}: Comprehensive evaluation of the tightly coupled multi-camera LIVO framework on public benchmarks and custom datasets.
\end{enumerate}

\section{Related Works}

\subsection{LiDAR-Inertial Odometry Evolution}
LiDAR-inertial odometry (LIO) systems evolved from feature-based LiDAR odometry like LOAM \cite{zhang2014loam} and LeGO-LOAM \cite{shan2018lego} to direct methods. FAST-LIO \cite{xu2021fast} eliminated feature extraction and directly registered raw points using an iterated Kalman filter. FAST-LIO2 \cite{xu2022fast} further improved efficiency through incremental map updates with ikd-Tree. However, these systems lack visual information for operation in geometrically degraded environments.

\subsection{LiDAR-Visual-Inertial Integration}
LiDAR-visual-inertial fusion addresses pure LIO limitations. R$^2$LIVE \cite{lin2021r} employs sparse visual features with re-projection error minimization. R$^3$LIVE \cite{lin2022r} advanced to direct photometric alignment with RGB-color consistency. FAST-LIVO \cite{zheng2022fast} pioneered tight coupling by attaching image patches to LiDAR map points. FAST-LIVO2 \cite{zheng2024fast} refined this through sequential ESIKF updates with direct methods. However, existing LIVO systems predominantly employ single-camera configurations, limiting comprehensive exploitation of wide-FoV LiDAR geometry across extended spatial coverage.

\subsection{Multi-Camera Visual-Inertial Systems}
Multi-camera visual-inertial SLAM has gained attention for extended spatial coverage. MIMC-VINS \cite{eckenhoff2021mimc} supports arbitrary camera numbers using multi-state constraint Kalman filtering. BAMF-SLAM \cite{zhang2023bamf} integrates multiple fisheye cameras with IMU through bundle adjustment. For large FoV coverage, LF-VIO \cite{wang2022lfvio} and LF-VISLAM \cite{wang2023lfvislam} employ negative imaging plane models for single wide-FoV cameras. MAVIS \cite{wang2024mavis} extends to partially overlapped multi-camera systems. These systems operate in the visual-inertial domain without leveraging LiDAR geometric structure.

\subsection{Direct Visual Methods}
Direct visual methods minimize photometric error without feature extraction. DTAM \cite{newcombe2011dtam}, LSD-SLAM \cite{engel2014lsd}, and DSO \cite{wang2017stereo} established foundations for dense and sparse tracking. MultiCol-SLAM \cite{urban2016multicol} extended direct methods to multi-camera systems. This suits multi-camera LIVO by avoiding correspondence complexity in non-overlapping views.

\section{System Overview And Methodology}

\begin{figure*}[!htbp]
    \centering
    \includegraphics[width=\textwidth]{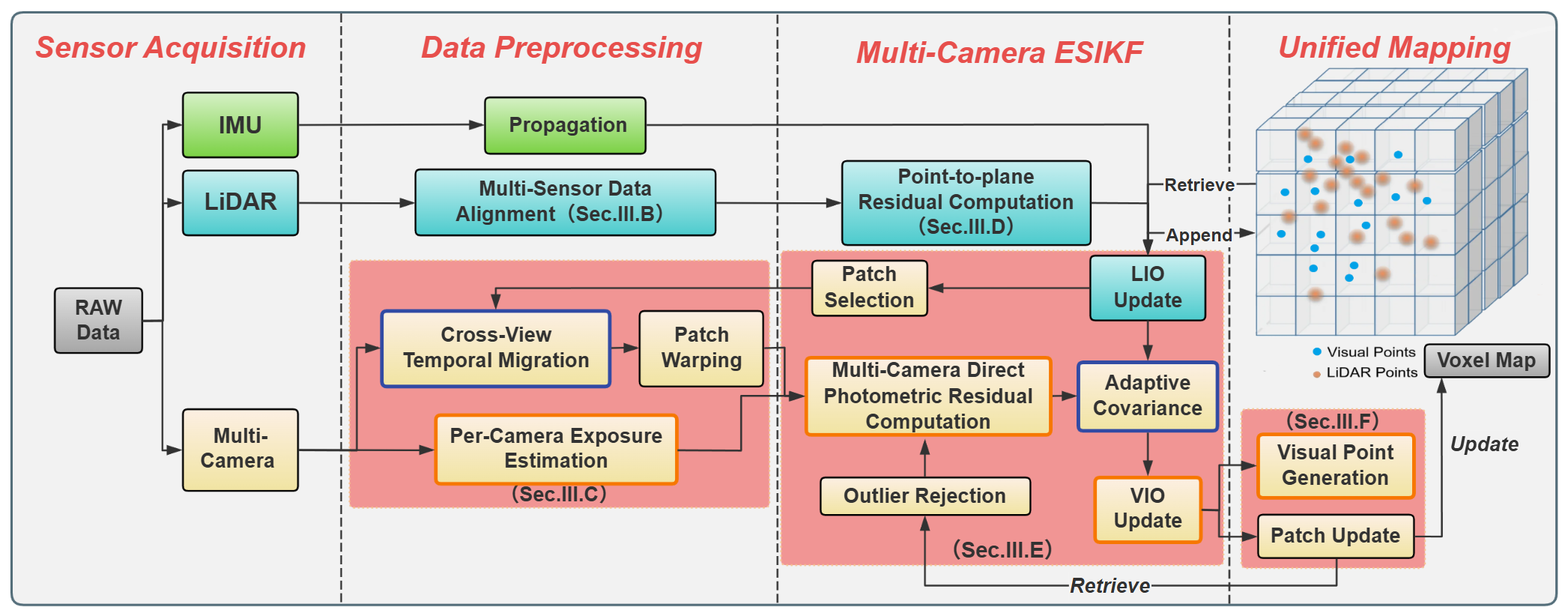}
    \caption{Overview of the Omni-LIVO architecture. Light yellow, light blue, and light green modules represent visual, LiDAR, and IMU data respectively. Orange borders indicate extensions to baseline methods; blue borders indicate novel contributions.}
    \label{fig:Omni-LIVO}
    \end{figure*}

The Omni-LIVO system comprises four modules: Sensor Acquisition, Preprocessing, Multi-Camera ESIKF, and Unified Mapping, as illustrated in Fig.~\ref{fig:Omni-LIVO}. The processing pipeline begins with IMU propagation, followed by sequential ESIKF updates that integrate LiDAR point-to-plane residuals with multi-view photometric constraints. Cross-View temporal migration and adaptive covariance preserve photometric consistency across non-overlapping views.

\subsection{Notational Conventions}

Table~\ref{table:notation} summarizes the key mathematical notation used throughout this paper.

\begin{table}[!t]
\caption{Notational Conventions\label{table:notation}}
\centering
\small
\renewcommand{\arraystretch}{0.8}
\setlength{\tabcolsep}{3pt}
\begin{tabular}{cl}
\toprule
\textbf{Symbol} & \textbf{Description} \\
\midrule
$\mathbf{T}^w_c$ & Transformation from world to camera frame \\
$\mathbf{R}$, $\mathbf{t}$ & Rotation matrix and translation vector \\
$\mathbf{p}$ & 3D point in world coordinates \\
$\tau_c$ & Inverse exposure time parameter for camera $c$ \\
$N_c$ & Number of cameras \\
$\mathcal{P}_k^{\text{seg}}$ & Temporally segmented LiDAR point cloud \\
$\mathcal{C}_k^{(i)}$ & Image frame from camera $i$ at time $k$ \\
$\mathcal{V}_k$ & Voxel containing geometric and visual data \\
$\mathbf{n}_k$ & Plane normal vector \\
$\mathbf{r}_{VIO}$ & Visual-inertial odometry residual vector \\
$\mathbf{r}_{LIO}$ & LiDAR-inertial odometry residual \\
$\hat{\mathbf{x}}^-$, $\mathbf{P}^-$ & Prior state estimate and covariance \\
$\hat{\mathbf{x}}_{k|k}$, $\mathbf{P}_{k|k}$ & Posterior state estimate and covariance \\
$\mathbf{H}$ & Jacobian matrix \\
$\mathbf{K}$ & Kalman gain \\
$\ell$ & Iteration index \\
$\boxplus$ & Manifold addition operator \\
\bottomrule
\end{tabular}
\end{table}

\subsection{Multi-Sensor Data Alignment}
Building upon FAST-LIVO2 \cite{zheng2024fast}, we extend temporal alignment to multi-camera configurations \cite{nguyen2021miliom}. The system forms temporally coherent measurement packets:

\begin{equation}
\label{eq1}
\mathcal{M}_{\mathrm{k}} = \{\mathcal{P}_{\mathrm{k}}^{\text{seg}}, \{\mathcal{J}_{\mathrm{j}}\}_{\mathrm{j}=\mathrm{k}-1}^{\mathrm{k}}, \{\mathcal{C}_{\mathrm{k}}^{(\mathrm{i})}\}_{\mathrm{i}=1}^{N_c}\}
\end{equation}

where $\mathcal{M}_k$ denotes the multi-sensor measurement packet at time $k$, and $\{\mathcal{J}_{\mathrm{j}}\}_{\mathrm{j=k-1}}^{\mathrm{k}}$ represents the set of high-frequency IMU measurements spanning the inter-frame interval.

Although LiDAR and cameras may operate at the same nominal rate (e.g., 10 Hz), their trigger events are not always aligned. To align data across modalities, each LiDAR point $p_i$ with timestamp $\tau_i$ is assigned to the temporally nearest camera frame:

\begin{equation}
\label{eq2}
p_i\to\text{frame}_k, \quad \text{with } k=\operatorname*{arg\,min}_j|\tau_i-t_\mathrm{sync}^{(j)}|
\end{equation}
where $t_\mathrm{sync}^{(j)}$ is the sync trigger time. Within each frame $k$, all camera channels are externally triggered to ensure simultaneous exposure:
\begin{equation}
\label{eq3}
t_\mathrm{cam}^{(i,k)}=t_\mathrm{sync}^{(k)}, \quad \forall i \in [1, N_c]
\end{equation}
where $t_\mathrm{cam}^{(i,k)}$ is the trigger time for camera $i$ at frame $k$.

\begin{figure}[!t]
\centering
\includegraphics[width=\columnwidth]{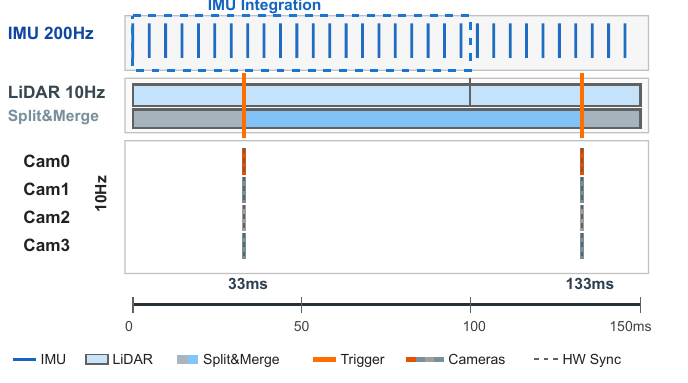}%
\caption{Multi-sensor temporal alignment. LiDAR points are associated with the nearest camera frame, while IMU measurements provide continuous motion priors for temporally consistent fusion.}
\label{fig:sync}
\end{figure}
\subsection{Cross-View Direct Method}
\begin{figure}[!t]
\centering
\includegraphics[width=\columnwidth]{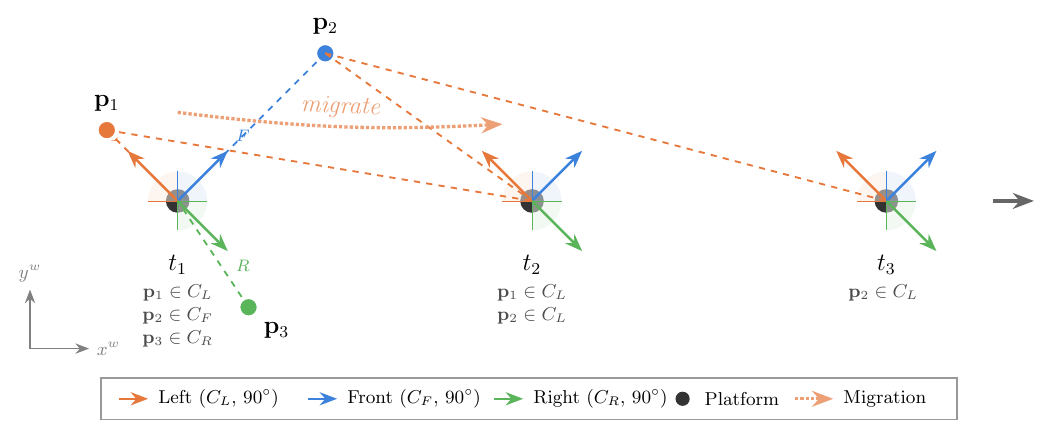}
\caption{Cross-View temporal migration. At $t_1$, map points $\mathbf{p}_1$, $\mathbf{p}_2$, $\mathbf{p}_3$ are observed by Left ($C_L$), Front ($C_F$), and Right ($C_R$) cameras respectively. As the platform moves, points migrate between camera views ($\mathbf{p}_2$ migrates from $C_F$ to $C_L$ at $t_2$), maintaining photometric continuity across non-overlapping $90^\circ$ FoVs through Cross-View residuals.}
\label{fig:migration}
\end{figure}

We formalize Cross-View temporal migration to maintain photometric continuity across non-overlapping views (Fig.~\ref{fig:migration}). The migration indicator $M(\mathbf{p}, i \rightarrow j)$ determines whether a map point transitions from camera $i$ to $j$:
\begin{equation}
\label{eq4}
M(\mathbf{p}, \mathrm{i} \rightarrow \mathrm{j}) = 1 \Leftrightarrow
\begin{cases}
\text{cam}_{\text{ref}}(\mathbf{p}) = \mathrm{i} \\
\text{cam}_{\text{cur}}(\mathbf{p}) = \mathrm{j} \\
\mathrm{i} \neq \mathrm{j} \\
\text{exclusive}_{\text{visible}}(\mathbf{p}, \mathrm{j}) = 1 \\
\text{temporal}_{\text{migrated}}(\mathbf{p}, \mathrm{i} \rightarrow \mathrm{j}) = 1
\end{cases}
\end{equation}
where $\text{cam}_{\text{ref}}(\mathbf{p})$ and $\text{cam}_{\text{cur}}(\mathbf{p})$ return the camera IDs observing $\mathbf{p}$ in reference and current frames, $\text{exclusive}_{\text{visible}}(\mathbf{p}, \mathrm{j}) = 1$ ensures $\mathbf{p}$ is visible only in camera $j$ (i.e., $\pi_j(\mathbf{T}_j^w \cdot \mathbf{p}) \in \Omega_j$ with $\Omega_j$ being the valid image region, and invisible in other cameras), and $\text{temporal}_{\text{migrated}}(\mathbf{p}, \mathrm{i} \rightarrow \mathrm{j}) = 1$ verifies $\mathbf{p}$ transitioned from camera $i$ to $j$ across successive frames.

When camera migration occurs, the reference patch is dynamically switched by prioritizing observations from the current camera, or selecting the patch with minimum photometric error:
\begin{equation}
\label{eq4b}
\mathbf{p}_{\mathrm{ref}}^* = \operatorname*{arg\,min}_{\mathbf{p}_k \in \mathcal{O}(\mathbf{p})} \frac{1}{|\mathcal{O}(\mathbf{p})|-1} \sum_{\mathbf{p}_m \in \mathcal{O}(\mathbf{p}) \setminus \{\mathbf{p}_k\}} \|\mathbf{I}_k - \mathbf{I}_m\|^2
\end{equation}
where $\mathcal{O}(\mathbf{p})$ denotes the set of all observations of point $\mathbf{p}$, $\mathbf{I}_k \in \mathbb{R}^{P}$ is the vectorized patch intensity, and $P$ is the patch size.

To ensure consistent intensity comparison across cameras, image measurements are normalized by vignetting:

\begin{equation}
\label{eq5}
I^\prime_k(u,v)=V_{\mathrm{k}}(r^2) \cdot I_{\mathrm{k}}(u,v)
\end{equation}
where $I_k(u,v)$ denotes the raw image intensity at pixel coordinates $(u,v)$, $I^\prime_k(u,v)$ is the vignetting-normalized intensity, $r$ is the radial distance from the image center, and $V_k(r^2)$ is a vignetting function modeled as:

\begin{equation}
\label{eq6}
V_k(r^2)=1+\sum^3_{n=1}\beta_{k,n}\cdot r^{2n}
\end{equation}
where $\beta_{k,n}$ are the vignetting polynomial coefficients calibrated offline for each camera.

Exposure variations across cameras are handled through per-camera inverse exposure time parameters $\tau_c$ jointly optimized in the ESIKF framework, ensuring photometric consistency without requiring relative exposure factor estimation.

The relative transformation and induced homography under local planar assumption are formulated to warp patches between cameras. The relative transformation from camera $i$ to camera $j$ is:
\begin{equation}
\label{eq7}
\mathbf{T}_{\mathrm{i} \to \mathrm{j}}(t) = \mathbf{T}^{\mathrm{w}}_{\mathrm{j}}(t_{\mathrm{j}}) \cdot (\mathbf{T}^{\mathrm{w}}_{\mathrm{i}}(t_{\mathrm{i}}))^{-1}
\end{equation}

Under the local planar assumption, the homography matrix $\mathbf{H}_{i \rightarrow j}(\mathbf{p})$ that maps points from camera $i$ to camera $j$ is \cite{hartley2003multiple}:
\begin{equation}
\label{eq8}
\mathbf{H}_{\mathrm{i} \rightarrow \mathrm{j}}(\mathbf{p}) = \mathbf{R}_{\mathrm{i} \rightarrow \mathrm{j}} + \frac{\mathbf{t}_{\mathrm{i} \rightarrow \mathrm{j}} \mathbf{n}_{\mathrm{i}}^{\mathrm{T}}}{d_{\mathrm{i}}}
\end{equation}
where $\mathbf{R}_{\mathrm{i} \rightarrow \mathrm{j}}$ and $\mathbf{t}_{\mathrm{i} \rightarrow \mathrm{j}}$ are the rotation and translation from $\mathbf{T}_{\mathrm{i} \to \mathrm{j}}$, $\mathbf{n}_{\mathrm{i}}$ is the unit surface normal at the patch in camera $i$ frame, and $d_{\mathrm{i}}$ is the point depth in camera $i$ frame. This homography enables direct patch warping across non-overlapping camera views for photometric alignment.

\begin{figure}[!t]
\centering
\includegraphics[width=1\columnwidth]{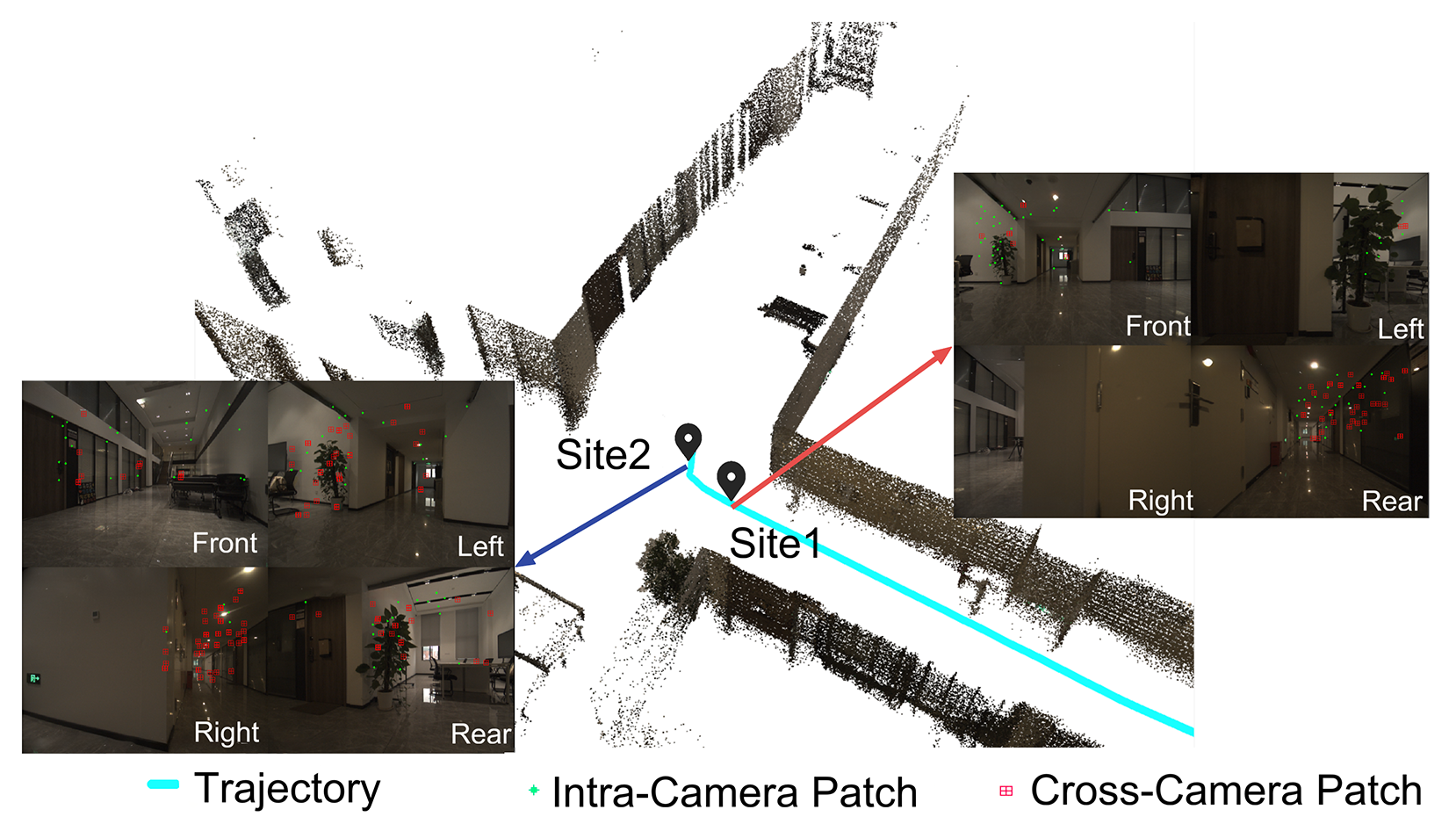}%
\caption{Photometric patch associations during turning: intra-camera patches (green) remain within the same view, while migrated patches (red) form Cross-View associations across different cameras.}
\label{fig:associations}
\end{figure}

Based on these definitions, we formulate both intra-camera and Cross-View residuals, with the patch association process visualized in Fig.~\ref{fig:associations}. For intra-camera alignment, the direct photometric residual $r_{\text{direct}}$ is:
\begin{equation}
\label{eq9}
r_{\text{direct}}^{\mathrm{c}}(\mathbf{p}) = \tau_{c} \cdot I_{c}^{\prime} ( \pi_{c} ( \mathbf{T}_{c}^{w} \cdot \mathbf{p} ) ) - \tau_{\mathrm{ref}} \cdot I_{\mathrm{ref}}^{\prime} ( \pi_{\mathrm{ref}} ( \mathbf{p}_{\mathrm{ref}} ) )
\end{equation}
where $\pi_c(\cdot)$ denotes the camera projection function from 3D to 2D image coordinates, and subscript ``ref'' denotes the reference frame.

Cross-View temporal migration residuals $r_{\mathrm{migration}}$ establish constraints between different cameras across time:
\begin{equation}
\label{eq10}
r_{\mathrm{migration}}^{i \to j} ( p ) = \tau_{i} \cdot W_{H} \left[ I^\prime_{i} , \pi_{i} ( p , t_{i} ) \right] - \tau_{j} \cdot I^\prime_{j} ( \pi_{j} ( p , t_{j} ) )
\end{equation}
where camera $i$ is the source (reference) and camera $j$ is the target (current), $W_H$ represents warping using homography $H$, and $t_i$, $t_j$ are the respective camera timestamps.

For cross-camera constraints, we formulate bidirectional coupled Jacobians that account for geometric contributions from both cameras:
\begin{equation}
\label{eq10b}
\mathbf{J}_{\mathbf{R}}^{\mathrm{coupled}} = \mathbf{J}_{\mathbf{R}}^{i} - \mathbf{J}_{\mathbf{R}}^{j}, \quad
\mathbf{J}_{\mathbf{t}}^{\mathrm{coupled}} = \mathbf{J}_{\mathbf{t}}^{i} - \mathbf{J}_{\mathbf{t}}^{j}
\end{equation}
where $\mathbf{J}_{\mathbf{R}}^{k} = \tau_k \frac{\partial I_k}{\partial \mathbf{R}}$ and $\mathbf{J}_{\mathbf{t}}^{k} = \tau_k \frac{\partial I_k}{\partial \mathbf{t}}$ are exposure-weighted Jacobians with respect to rotation and translation for camera $k$. Similarly, exposure Jacobians are computed as $\frac{\partial r^{i \to j}}{\partial \tau_i} = I_i$ and $\frac{\partial r^{i \to j}}{\partial \tau_j} = -I_j$. This bidirectional formulation enforces both geometric and photometric consistency between camera pairs.

The complete residual vector $\mathbf{r}_{VIO}$ stacks all intra-camera and Cross-View residuals for joint optimization:
\begin{equation}
\label{eq11}
\mathbf{r}_{VIO} = \left[ \mathbf{r}^{1}_{\mathrm{direct}}, \ldots, \mathbf{r}^{N_c}_{\mathrm{direct}}, \mathbf{r}^{1 \to 2}_{\mathrm{migration}}, \ldots, \mathbf{r}^{i \to j}_{\mathrm{migration}} \right]^T
\end{equation} 

Temporal migration constraints preserve photometric continuity across cameras, enhancing geometric diversity and pose estimation robustness. Multi-camera configurations enable stricter photometric thresholds while maintaining sufficient observations, improving both efficiency and accuracy.

\subsection{LiDAR Geometric Constraint Processing}

LiDAR measurements provide structural constraints that complement photometric observations in our multi-camera framework. We employ voxel-based plane extraction to efficiently process point clouds and generate geometric residuals.

The raw LiDAR point cloud is discretized into voxels of resolution $v_{\text{size}}$. For each voxel $\mathcal{V}_k$ containing $N_k$ points $\mathcal{P}_k = \{ p_i\}^{N_k}_{i=1}$, we compute the voxel centroid $\bar{p}_k = \frac{1}{N_k}\sum^{N_k}_{i=1} p_i$ and the sample covariance matrix $\mathbf{C}_k$:

\begin{equation}
\label{eq14}
\mathbf{C}_k=\frac{1}{N_k} \sum^{N_k}_{i=1}(p_i-\bar{p}_k)(p_i-\bar{p}_k)^{\mathrm{T}}
\end{equation}

Planar voxels are identified via eigen-decomposition of $\mathbf{C}_k$. If the smallest eigenvalue $\lambda_{\min} < \epsilon_{\text{plane}}$ (where $\epsilon_{\text{plane}}$ is the planarity threshold), the corresponding eigenvector defines the plane normal $\mathbf{n}_k$, which satisfies the plane equation for any point $p$ on the plane:

\begin{equation}
\label{eq15}
\mathbf{n}_k^{\mathrm{T}}(p-\bar{p}_k)=0
\end{equation}

LiDAR geometric residuals are constructed by measuring point-to-plane distances. Let $\mathbf{p}^{\mathrm{w}}$ denote a LiDAR point in world frame and $\mathbf{T}_{\mathrm{L}}^{\mathrm{w}}$ the world-to-LiDAR transformation. The point in LiDAR frame is $\mathbf{p}^{\mathrm{L}} = \mathbf{R}_{\mathrm{L}}^{\mathrm{w}} \mathbf{p}^{\mathrm{w}} + \mathbf{t}_{\mathrm{L}}^{\mathrm{w}}$, and the LiDAR-inertial residual $\mathbf{r}_{LIO}$ measuring point-to-plane distance is:

\begin{equation}
\label{eq16}
\mathbf{r}_{LIO} ( \mathbf{p}^{\mathrm{w}} ) = \mathbf{n}_{k}^{\mathrm{T}} ( \mathbf{R}_{\mathrm{L}}^{\mathrm{w}} \mathbf{p}^{\mathrm{w}} + \mathbf{t}_{\mathrm{L}}^{\mathrm{w}} - \bar{\mathbf{p}}_{k} )
\end{equation}
where $\bar{\mathbf{p}}_k$ is the plane centroid in LiDAR frame.

\subsection{Iterative Multi-Camera State Estimation with ESIKF}
\begin{figure*}[!t]
\centering
\includegraphics[width=1\textwidth]{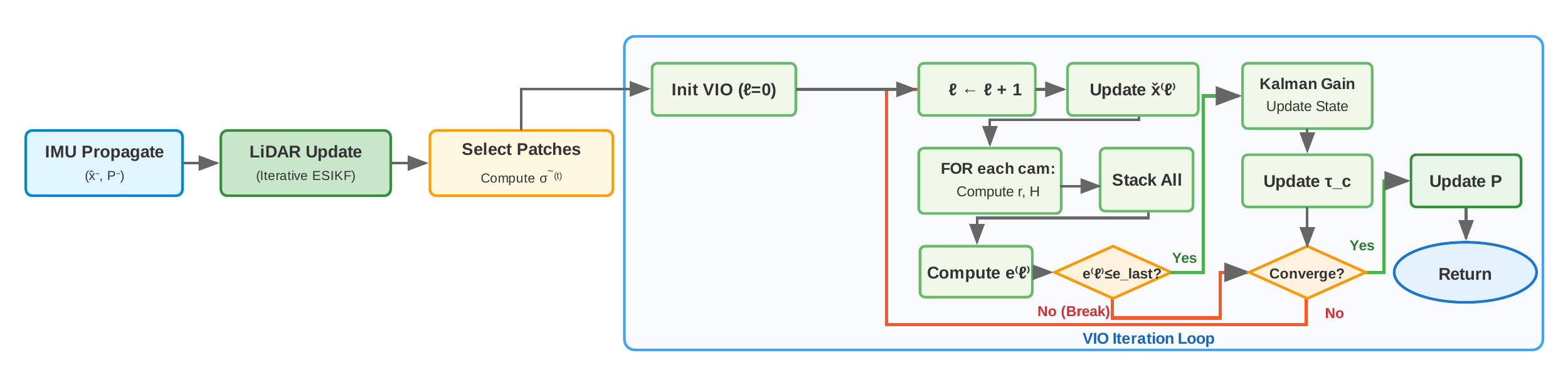}
\caption{Iterative Multi-Camera ESIKF Framework. The pipeline performs Synchronize , IMU Propagate , LiDAR Update, Select Patches, and Adaptive Cov,
followed by Init VIO and iterative updates of the photometric VIO loop until convergence.}
\label{fig:ESIKF}
\end{figure*}

Measurement quality varies across cameras due to heterogeneous exposure and environments. We extend ESIKF to integrate multi-view photometric constraints with LiDAR measurements through sequential update and adaptive covariance, as shown in Fig.~\ref{fig:ESIKF}. The state is first updated with LiDAR point-to-plane residuals, then sequentially updated with multi-camera photometric residuals. Multi-camera residuals and Jacobians are stacked across all cameras and processed through iterative ESIKF updates until convergence.

For heterogeneous exposure settings, independent inverse exposure time parameters $\tau_c$ are maintained for each camera $c \in [1, N_c]$ and jointly optimized within the ESIKF framework. The Jacobian matrix is extended to $\mathbb{R}^{m \times (6+N_c)}$, where $m$ is the total number of measurements, 6 represents pose parameters (rotation and translation), and $N_c$ represents per-camera exposure parameters.

The photometric covariance matrix $R_{\mathrm{img}}$ is adaptively scaled based on real-time photometric error to account for varying measurement quality. The covariance scaling factor $\sigma^{(t)}$ at time $t$ is computed via linear mapping:
\begin{equation}
\label{eq19}
\sigma^{(t)} = \sigma_{\min} + \frac{\bar{e}^{(t-1)} - 1}{e_{\max} - 1} \cdot (\sigma_{\max} - \sigma_{\min})
\end{equation}
where $\bar{e}^{(t-1)}$ is the mean squared photometric error from the previous frame, $\sigma_{\min}$ and $\sigma_{\max}$ are the minimum and maximum covariance scaling bounds (set to $100.0$ and $1000.0$ respectively), and $e_{\max}$ is the maximum error threshold for scaling (typically set to $100.0$). Temporal smoothing ensures stability:
\begin{equation}
\label{eq20}
\tilde{\sigma}^{(t)} = \lambda \cdot \sigma^{(t)} + (1-\lambda) \cdot \tilde{\sigma}^{(t-1)}
\end{equation}
where $\tilde{\sigma}^{(t)}$ is the smoothed covariance scaling factor and $\lambda$ is the smoothing parameter (set to $0.3$). This adaptive weighting down-weights cameras with high photometric errors (e.g., motion blur, occlusion) and up-weights reliable observations, improving robustness in adverse conditions.

Algorithm~\ref{alg:multicam_livo} summarizes the complete implementation. Algorithm-specific notation: $\check{\mathbf{x}}^{(\ell)}$ is the linearization point; $\mathbf{r}_c^{(\ell)}$, $\mathbf{H}_c^{(\ell)}$ are per-camera residuals and Jacobians stacked into $\mathbf{r}^{(\ell)}_{VIO}$, $\mathbf{H}^{(\ell)}_{VIO}$; $\delta\hat{\mathbf{x}}^{(\ell)}$ is the state increment; $e^{(\ell)}$ is the mean squared error; $\ell_{\max}=10$ is the maximum iteration count.
\begin{algorithm}
\small
\setlength{\algoheightrule}{0pt}
\SetAlCapSkip{0ex}
\SetAlgoSkip{}
\setlength{\baselineskip}{7pt}
\setlength{\lineskip}{0pt}
\setlength{\lineskiplimit}{0pt}
\setlength{\abovedisplayskip}{2pt}
\setlength{\belowdisplayskip}{2pt}
\SetInd{0.3em}{0.5em}
\SetAlgoVlined
\caption{Multi-Camera ESIKF Update}
\label{alg:multicam_livo}
\DontPrintSemicolon
\SetNlSty{tiny}{}{}
IMU propagation: $\hat{\mathbf{x}}^-$, $\mathbf{P}^-$; LiDAR plane extraction\;
\textbf{LiDAR update}: Iterative ESIKF with $\mathbf{r}_{LIO}$, $\mathbf{H}_{LIO}$\;
Select patches from cameras $c \in [1,N_c]$; Compute $\tilde{\sigma}^{(t)}$ via Eq.~(\ref{eq19})-(\ref{eq20})\;
\textbf{VIO update}: $\ell \leftarrow 0$, $\delta\hat{\mathbf{x}}^{(0)} \leftarrow \mathbf{0}$, $e_{\text{last}} \leftarrow \infty$\;
\Repeat{$\|\delta\mathbf{R}^{(\ell)}\| < 0.001^\circ$ and $\|\delta\mathbf{t}^{(\ell)}\| < 0.01$mm or $\ell \geq \ell_{\max}$}{
    $\ell \leftarrow \ell + 1$; Update $\check{\mathbf{x}}^{(\ell)} = \hat{\mathbf{x}}^- \boxplus \delta\hat{\mathbf{x}}^{(\ell-1)}$\;
    \For{camera $c = 1, \ldots, N_c$}{
        Compute $\mathbf{r}_c^{(\ell)}$, $\mathbf{H}_c^{(\ell)}$ with cross-camera constraints\;
    }
    Stack $\mathbf{r}^{(\ell)}_{VIO} = [\mathbf{r}_1^{(\ell)}; \ldots; \mathbf{r}_{N_c}^{(\ell)}]$, $\mathbf{H}^{(\ell)}_{VIO} = [\mathbf{H}_1^{(\ell)}; \ldots; \mathbf{H}_{N_c}^{(\ell)}]$\;
    Compute $e^{(\ell)} = \|\mathbf{r}^{(\ell)}_{VIO}\|^2 / n_{\text{meas}}$\;
    \eIf{$e^{(\ell)} \leq e_{\text{last}}$}{
        $e_{\text{last}} \leftarrow e^{(\ell)}$\;
        $\mathbf{K} = ((\mathbf{H}^{(\ell)}_{VIO})^{\mathrm{T}}\mathbf{H}^{(\ell)}_{VIO} + \mathbf{P}^{-1}/\tilde{\sigma}^{(t)})^{-1}$\;
        $\delta\hat{\mathbf{x}}^{(\ell)} = \mathbf{K}((\mathbf{H}^{(\ell)}_{VIO})^{\mathrm{T}}\mathbf{r}^{(\ell)}_{VIO} - \mathbf{P}^{-1}\delta\hat{\mathbf{x}}^{(\ell-1)}/\tilde{\sigma}^{(t)})$\;
        $\hat{\mathbf{x}}^- \leftarrow \hat{\mathbf{x}}^- \boxplus \delta\hat{\mathbf{x}}^{(\ell)}$\;
    }{
        \textbf{break}\;
    }
}
$\mathbf{P}_{k|k} = (\mathbf{I} - \mathbf{K}\mathbf{H}^{(\ell)}_{VIO}) \mathbf{P}^-$\;
\Return{$\hat{\mathbf{x}}_{k|k}$, $\mathbf{P}_{k|k}$}
\end{algorithm}

\subsection{Unified Voxel Mapping with Multi-View Integration}
We integrate LiDAR geometry and multi-view photometric patches within a single adaptive volumetric structure employing density-driven octree refinement \cite{yuan2022efficient}. Each voxel $\mathcal{V}_k$ encapsulates three components:
\begin{equation}
\label{eq17}
\mathcal{V}_k = \{ \mathcal{G}_k, \mathcal{W}_k, \mathcal{S}_k \}
\end{equation}
where $\mathcal{G}_k$ represents geometric plane parameters from LiDAR, $\mathcal{W}_k$ contains visual map points with multi-view observations, and $\mathcal{S}_k$ stores octree subdivision structure. Visual map points are generated by partitioning images into regular grids and selecting LiDAR points with highest Shi-Tomasi corner response. Each visual point $j$ maintains a feature structure $\mathcal{F}_j$ containing $N_{\text{obs}}$ observations across multiple frames:
\begin{equation}
\label{eq18}
\mathcal{F}_j = \{ \mathcal{P}^{(k)}_j, \mathrm{T}^{(k)}_{wc}, \tau^{(k)}, id^{(k)}_{\text{cam}} \}^{N_{\text{obs}}}_{k=1}
\end{equation}
where $N_{\text{obs}}$ is the number of observations, $\mathcal{P}^{(k)}_j$ is the patch organized in multi-level pyramid structure, $\mathrm{T}^{(k)}_{wc}$ is the camera pose, $\tau^{(k)}$ is the exposure parameter, and $id^{(k)}_{\text{cam}}$ is the camera identifier.

\section{Experimental Evaluation}

We evaluate the proposed system on public benchmarks and custom datasets, and compare with state-of-the-art methods.

\subsection{Datasets and Baselines}

To demonstrate the effectiveness of the proposed multi-camera LIVO framework, we conduct comprehensive experiments evaluating trajectory accuracy, ablation of key components, computational efficiency, and qualitative mapping performance across diverse environments.

\begin{figure}[!htbp]
\centering
\subfloat[Handheld device setup\label{fig:handheld}]{
    \includegraphics[width=0.45\columnwidth]{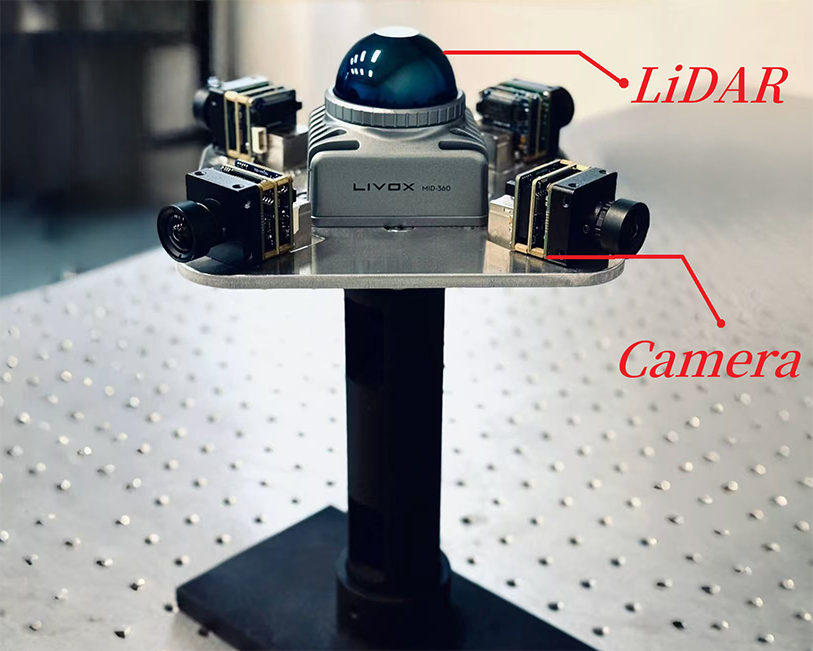}
}
\hfill
\subfloat[Robotic platform setup\label{fig:robot}]{
    \includegraphics[width=0.45\columnwidth]{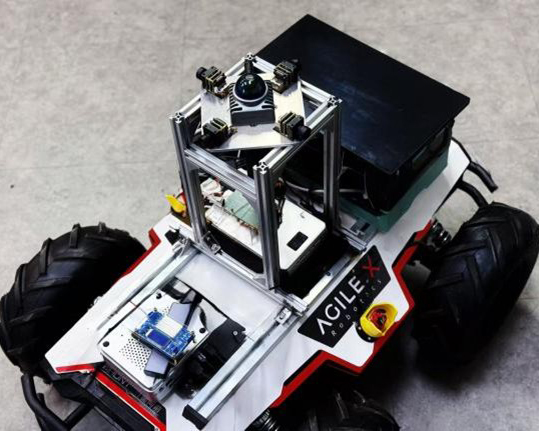}
}
\caption{SLAM data acquisition platforms used in our custom dataset. (a) Handheld configuration. (b) Robotic configuration.}
\label{fig:dataset_platforms}
\end{figure}

Evaluation is conducted on Hilti SLAM Challenge 2022/2023 and Newer College Dataset. We additionally collected a custom dataset comprising multiple sequences using LIVOX MID360 LiDAR (360° horizontal FoV, 10Hz) and four cameras (1024$\times$768, 10Hz) in cross-pattern configuration with handheld and robotic platforms (Fig.~\ref{fig:dataset_platforms}).

\begin{table*}[!htbp]
    \caption{Absolute Trajectory Errors (RMSE in meters) on Multiple Benchmark Datasets}
    \label{table:RMSE}
    \centering
    \small
    \renewcommand{\arraystretch}{0.80}
    \setlength{\tabcolsep}{3.0pt}
    \begin{tabular}{cl|ccccc|ccccc}
    \toprule
    \multirow{2}{*}{\textbf{Dataset}} & \multirow{2}{*}{\textbf{Sequence}} & \multicolumn{5}{c|}{\textbf{Baseline Methods}} & \multicolumn{5}{c}{\textbf{Omni-LIVO Variants}} \\
    \cmidrule(lr){3-7} \cmidrule(lr){8-12}
    & & \scriptsize{\textbf{FAST-LIVO}} & \scriptsize{\textbf{FAST-LIVO2}} & \scriptsize{\textbf{R$^3$LIVE}} & \scriptsize{\textbf{FAST-LIO2}} & \scriptsize{\textbf{OpenMAVIS}} & \scriptsize{\textit{w/o ac}} & \scriptsize{\textit{w/o cvtm}} & \scriptsize{\textbf{1-Cam}} & \scriptsize{\textbf{2-Cam}} & \scriptsize{\textbf{Ours}} \\
    \midrule
    \multirow{15}{*}{\rotatebox{90}{Hilti 2022}}
    & Attic to Upper Gallery & 1.650& 0.150& $\times$ & $\times$ & 0.103& \underline{0.082}& 0.089& 0.176& 0.085& \textbf{0.069}\\
    & Attic to Upper Gallery 2 & 1.165& 0.123& $\times$ & 3.591& 0.255& 0.129& 0.141& \underline{0.122}& 0.134& \textbf{0.120} \\
    & Basement 2 & 0.129& 0.043& 0.097 & 0.136 & 0.082& 0.038& \underline{0.036}& 0.049& \textbf{0.031}& \textbf{0.031}\\
    & Construction Ground & 0.016& \textbf{0.010}& 0.015& 0.018& 0.094& \textbf{0.010}& \underline{0.011}& \underline{0.011}& \textbf{0.010}& \underline{0.011}\\
    & Construction Multilevel & 0.155& \textbf{0.020}& \underline{0.021}& 0.038& 0.933& \underline{0.021}& \textbf{0.020}& 0.022& \textbf{0.020}& \textbf{0.020}\\
    & Construction Stairs & 0.291& 0.039& 0.611& 0.625& 0.054& \textbf{0.028}& 0.032& 0.042& 0.031& \underline{0.030}\\
    & Construction Upper 1 & 0.032& 0.030& \underline{0.028}& 0.124 & 0.083& 0.029& 0.029& 0.029& 0.029& \textbf{0.027} \\
    & Construction Upper 2 & 0.089& 0.041& 0.041& 0.100& 0.040& \underline{0.019}& 0.020& 0.037& 0.022& \textbf{0.009}\\
    & Construction Upper 3 & 0.091& 0.045& 0.049& $\times$ & 0.075& 0.027& \underline{0.025}& 0.047& \underline{0.025}& \textbf{0.011}\\
    & Cupola & 0.270& 0.130& $\times$ & $\times$& 0.140& 0.133& \underline{0.129}& 0.144& 0.131& \textbf{0.126}\\
    & Cupola 2 & $\times$& 0.185& $\times$ & $\times$ & 0.136& 0.127& 0.146& 0.187& \underline{0.121}& \textbf{0.098}\\
    & Long Corridor & 0.072& 0.056& 0.069& 0.077& 0.098& \underline{0.052}& 0.055& 0.056& 0.053& \textbf{0.051}\\
    & Lower Gallery & 0.065& 0.033& 0.010& 0.056& 0.064& \underline{0.009}& 0.011& 0.021& 0.011& \textbf{0.006}\\
    & Lower Gallery 2& 0.387& 0.155& $\times$ & $\times$ & 0.178& 0.109& 0.119& 0.158& \underline{0.108}& \textbf{0.083}\\
    & Outside Building & 0.047& 0.031& 0.033& 0.029& 0.425& 0.020& \underline{0.018}& 0.022& \textbf{0.016}& \textbf{0.016}\\
    \midrule
    \multirow{5}{*}{\rotatebox{90}{Hilti 2023}}
    & Floor 0 & 0.030& 0.023& 0.045& 0.068& 0.060& \underline{0.021}& \textbf{0.020}& 0.030& 0.024& \textbf{0.020}\\
    & Floor 1 & 0.029& 0.020& 0.045& 0.055& 0.127& \underline{0.018}& 0.019& 0.020& 0.019& \textbf{0.017}\\
    & Floor 2 & 0.093& 0.022& 0.121& 0.121& 0.178& \textbf{0.012}& 0.022& 0.020& \underline{0.018}& \textbf{0.012}\\
    & Basement & 0.072& \underline{0.019}& 0.070& 0.190& 0.246& 0.021& 0.020& 0.020& \textbf{0.016}& \textbf{0.016}\\
    & Stairs & 0.230& 0.020& 0.202& 0.154& 0.225& \underline{0.010}& 0.013& 0.019& 0.013& \textbf{0.008}\\
    \midrule
    \multirow{7}{*}{\rotatebox{90}{Newer College}}
    & Quad easy & 0.099& 0.082& 0.086& 0.092& 0.084& \underline{0.077}& 0.081& 0.080& \underline{0.077}& \textbf{0.071} \\
    & Quad medium & 1.530& 0.073& \textbf{0.061}& 0.101 & 0.113 & 0.069& 0.074& 0.075& 0.070& \underline{0.065} \\
    & Quad hard & 1.757& 0.083& \underline{0.071}& 0.146 & 0.124 & 0.073& 0.075& 0.075& 0.072& \textbf{0.057} \\
    & Stairs & 0.124& 0.057& 0.224& $\times$ & $\times$ & 0.049& 0.051& \textbf{0.046}& 0.061& \underline{0.048}\\
    & Underground easy & 0.125& \underline{0.049}& \textbf{0.037}& 0.156 & 0.120 & 0.051& 0.050& \underline{0.049}& 0.051& \underline{0.049} \\
    & Underground medium & 0.088& 0.043& 0.047 & 0.153 & 0.130 & \underline{0.037}& 0.041& 0.045& 0.039& \textbf{0.034} \\
    & Underground hard & 0.157& 0.241& 0.069& 0.141 & 0.255 & 0.068& 0.072& 0.096& \underline{0.061}& \textbf{0.055} \\
    \bottomrule
    \end{tabular}
    \end{table*}
We compare against five methods: FAST-LIVO2, FAST-LIVO, R$^3$LIVE (LiDAR-inertial-visual), FAST-LIO2 (LiDAR-inertial), and OpenMAVIS (multi-camera visual-inertial). All LIVO systems use 10Hz for both camera and LiDAR data acquisition. FAST-LIVO2 and our system use identical LIO and IMU parameters. Ablation studies are conducted using five system variants: \textit{w/o ac} (Adaptive Covariance disabled), \textit{w/o cvtm} (Cross-View Temporal Migration disabled), \textit{1-Cam}, \textit{2-Cam}, and \textit{Ours} (complete system with three cameras).

\subsection{Quantitative Results}

Table \ref{table:RMSE} summarizes absolute trajectory errors (bold: best, underline: second-best) across all sequences. Omni-LIVO variants achieve the lowest error in 25 out of 27 sequences and successfully complete all evaluated tracks, outperforming FAST-LIVO2, R$^3$LIVE, FAST-LIO2, and OpenMAVIS across diverse environments including construction sites, indoor corridors, and underground scenarios. Compared to FAST-LIVO2, our system achieves an average 30\% error reduction across all 27 sequences, with improvements up to 82\% on Lower Gallery.

Ablation analysis demonstrates that Cross-View Temporal Migration improves accuracy in sequences with large viewpoint changes (Cupola 2: 0.098~m vs. 0.146~m \textit{w/o cvtm}), while Adaptive Covariance enhances robustness under illumination variations (Lower Gallery 2: 0.083~m vs. 0.109~m \textit{w/o ac}). Camera number comparison shows progressive gains from 1-Cam to 3-Cam (Ours), with the largest improvements in complex environments (Underground hard: 0.096~m to 0.055~m). This trend indicates that multi-camera fusion provides greater benefits in geometrically challenging scenarios.

\FloatBarrier
\subsection{Computational Performance}

Computational complexity is $\mathcal{O}(N_p \cdot P \cdot L)$ for single-camera and $\mathcal{O}((N_p + N_{\text{cross}}) \cdot P \cdot L)$ for multi-camera, where $N_p$ is visual points, $P$ is patch size, $L$ is pyramid levels, and $N_{\text{cross}}$ is cross-camera observations.

We use NVIDIA Jetson AGX Orin with 12-core ARM CPU (2.2GHz) and 64GB RAM running Ubuntu 20.04. System parameters: LiDAR voxel size 1.0~m, surface filtering threshold 0.5~m, maximum 150 visual points per frame.

Table \ref{table:performance} presents processing time and memory consumption. Compared to FAST-LIVO2, Omni-LIVO introduces 1.16$\times$ overhead on Hilti (60.3~ms vs. 52.1~ms, 3 cameras) and 1.21$\times$ on custom sequences (40.6~ms vs. 33.6~ms, 4 cameras), primarily due to cross-camera constraint computation, while maintaining real-time 10Hz performance with 2906~MB (Hilti) and 2359~MB (Custom) memory consumption, demonstrating practical viability for embedded platforms.

\subsection{Qualitative Results}

\begin{figure*}[!htbp]
    \centering
    \subfloat[Cupola 2\label{fig:cupola2}]{%
        \includegraphics[width=\textwidth,height=0.30\textheight,keepaspectratio]{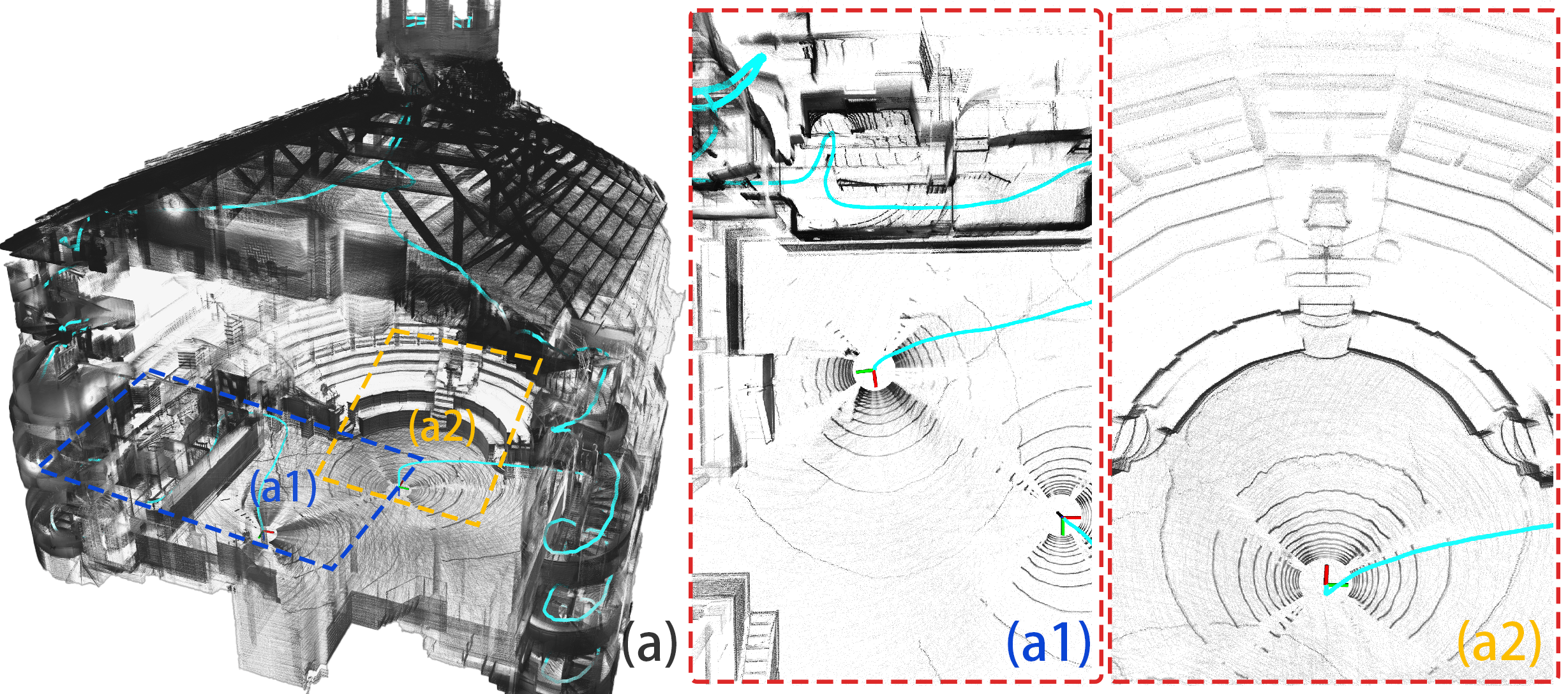}%
    }\\
    \subfloat[Underground Hard\label{fig:underground}]{%
        \includegraphics[width=\textwidth,height=0.20\textheight,keepaspectratio]{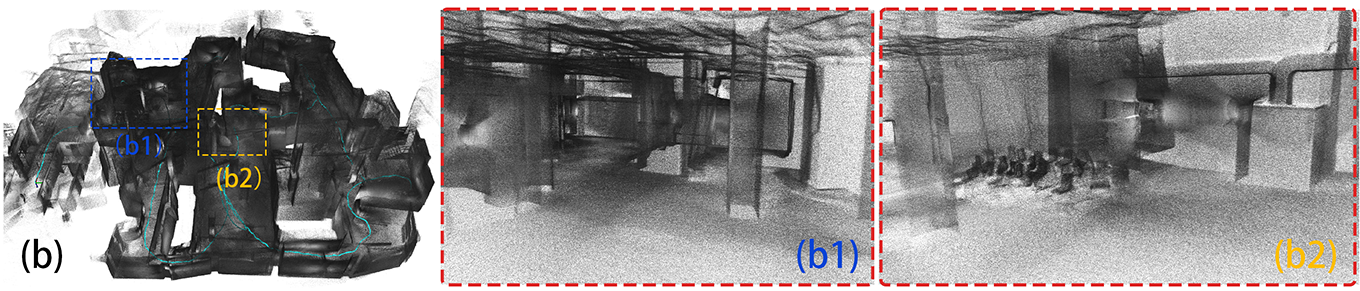}%
    }
    \caption{Mapping quality comparison on two challenging sequences. (a) Cupola 2 scene with overall map and detailed regions. (b) Underground Hard scene with overall map and detailed regions. The colored regions are grayscale LiDAR point clouds colorized using multi-camera gray images.}
    \label{fig:hilti_comparison}
\end{figure*}

\begin{figure*}[!htbp]
    \centering
    \subfloat[Overall map\label{fig:stairs_overall}]{\includegraphics[width=0.32\textwidth]{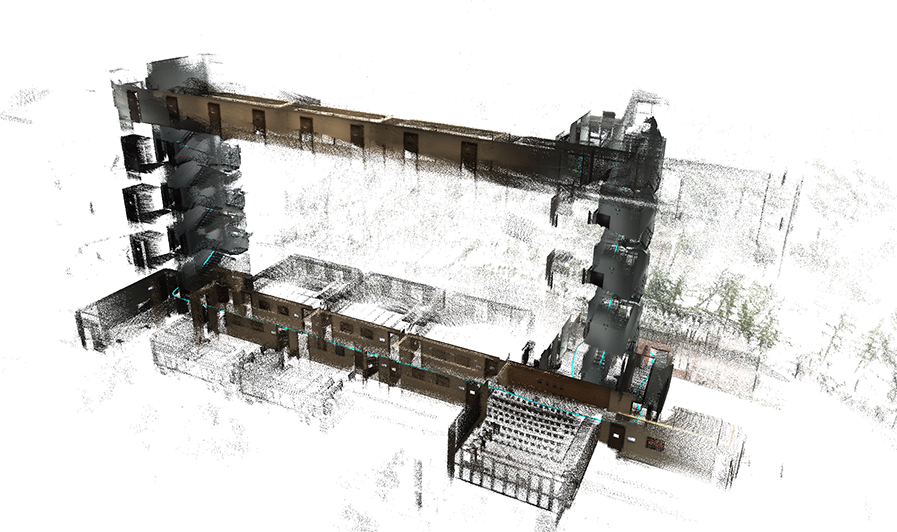}}%
    \hfil
    \subfloat[Omni-LIVO: Start/End alignment\label{fig:stairs_ours}]{\includegraphics[width=0.32\textwidth]{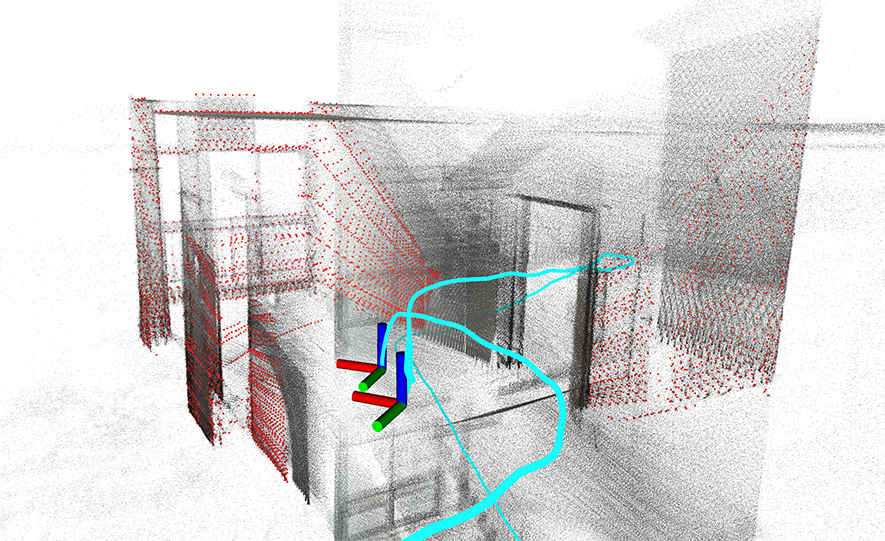}}%
    \hfil
    \subfloat[FAST-LIVO2: Start/End misalignment\label{fig:stairs_fastlivo2}]{\includegraphics[width=0.32\textwidth]{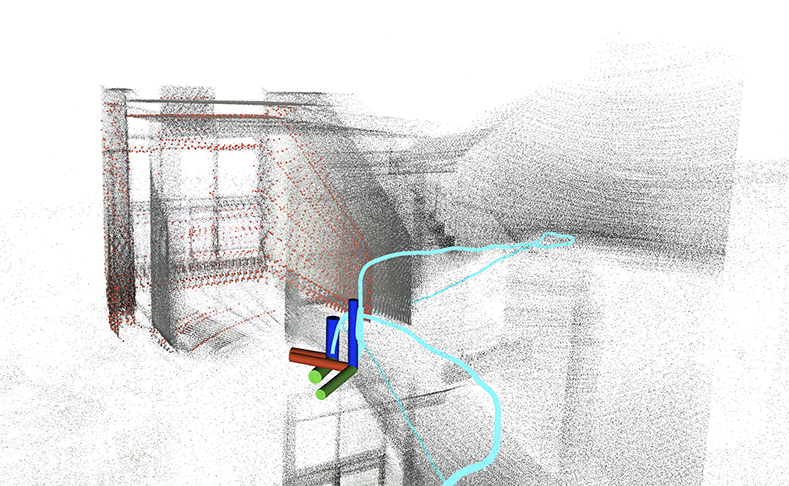}}%
    \caption{Loop closure comparison on staircase sequence. (a) Overall map. (b) Omni-LIVO: precise start/end alignment. (c) FAST-LIVO2: visible drift.}
    \label{fig:stairs_comparison}
\end{figure*}

Fig.~\ref{fig:hilti_comparison} demonstrates Omni-LIVO's mapping quality on 2 sequences with multi-camera configurations. Fig.~\ref{fig:stairs_comparison} validates the system's ability to maintain global consistency and achieve precise loop closure, with Omni-LIVO showing superior alignment compared to FAST-LIVO2's visible drift. The improved consistency is attributed to enhanced observability from multi-view constraints. Table~\ref{table:pointcloud_stats} shows that Omni-LIVO produces more RGB-colored points than competing methods, with particularly significant improvements in complex environments (3.5$\times$ more than FAST-LIVO2 in basement3).

\begin{table}[!htbp]
    \caption{Computational Performance Comparison\label{table:performance}}
    \centering
    \footnotesize
    \renewcommand{\arraystretch}{0.8}
    \setlength{\tabcolsep}{4pt}
    \begin{tabular}{lccccc}
    \toprule
    \multirow{2}{*}{\textbf{Sequence}} & \multirow{2}{*}{\textbf{\#Cam}} & \multicolumn{2}{c}{\textbf{FAST-LIVO2}} & \multicolumn{2}{c}{\textbf{Omni-LIVO}} \\
    \cmidrule(lr){3-4} \cmidrule(lr){5-6}
    & & \textbf{Time} & \textbf{Mem} & \textbf{Time} & \textbf{Mem} \\
    & & \textbf{(ms)} & \textbf{(MB)} & \textbf{(ms)} & \textbf{(MB)} \\
    \midrule
    Construction Upper & 3 & 50.7 & 1901 & 58.1 & 2017 \\
    Cupola 2 & 3 & 51.2 & 3477 & 57.0 & 3498 \\
    Outside Building & 3 & 54.5 & 3261 & 65.9 & 3203 \\
    \midrule
    \textit{Avg. (Hilti)} & 3 & 52.1 & 2880 & 60.3 & 2906 \\
    \midrule
    Corridor & 4 & 17.3 & 1054 & 21.0 & 1197 \\
    Basement3 & 4 & 49.9 & 3473 & 60.1 & 3520 \\
    \midrule
    \textit{Avg. (Custom)} & 4 & 33.6 & 2264 & 40.6 & 2359 \\
    \bottomrule
    \end{tabular}
    \end{table}
\begin{table}[!htbp]
\caption{RGB Point Cloud Statistics (Number of Colored Points)\label{table:pointcloud_stats}}
\centering
\footnotesize
\renewcommand{\arraystretch}{0.8}
\setlength{\tabcolsep}{4pt}
\begin{tabular}{lcccc}
\toprule
\textbf{Method} & \textbf{stairs} & \textbf{classroom} & \textbf{basement3} & \textbf{corridor} \\
\midrule
Omni-LIVO & \textbf{14,184,500} & \textbf{4,484,889} & \textbf{33,513,608} & \textbf{8,715,790} \\
FAST-LIVO2 & 5,539,700 & 3,111,504 & 9,519,125 & 4,863,001 \\
FAST-LIVO & 3,827,955 & 1,611,539 & 5,019,200 & 2,543,585 \\
R$^3$LIVE & 393,242 & 619,310 & 1,213,929 & 1,007,194 \\
\bottomrule
\end{tabular}
\end{table}

\subsection{Discussion}

Experimental results show that Omni-LIVO achieves improved accuracy over baseline methods across evaluated sequences with moderate computational overhead. The multi-camera configuration provides extended spatial coverage and denser scene colorization. Ablation studies demonstrate that Cross-View constraints improve performance in sequences with large viewpoint changes, while adaptive covariance enhances robustness under illumination variations.
\section{Conclusion}

This letter presents Omni-LIVO, a tightly coupled multi-camera LiDAR-inertial-visual odometry system that extends direct visual methods to multi-view configurations through Cross-View temporal migration and adaptive ESIKF integration.
Evaluation on public benchmarks and custom datasets shows that the system achieves lower trajectory errors compared to state-of-the-art LIVO, LIO, and visual-inertial methods. The multi-camera configuration provides extended spatial coverage and produces denser RGB-colored point clouds. The quantitative and qualitative improvements demonstrate the effectiveness of the proposed multi-camera fusion approach. Future work will explore online loop closure and investigate multi-sensor calibration refinement.

\bibliographystyle{IEEEtran}

\end{document}